\documentclass[journal]{IEEEtran}
\IEEEoverridecommandlockouts
\usepackage[normalem]{ulem}
\usepackage{cite}
\usepackage{amsmath,amssymb,amsfonts}
\usepackage{algorithmic}
\usepackage{graphicx}
\usepackage{tabularx}
\usepackage{listliketab}
\usepackage[center]{caption}
\graphicspath{ {./images/} }
\usepackage{textcomp}
\usepackage{xcolor}
\usepackage{subcaption}
\usepackage{colortbl}
\usepackage{url}
\usepackage[obeyFinal]{todonotes}
\def\BibTeX{{\rm B\kern-.05em{\sc i\kern-.025em b}\kern-.08em
    T\kern-.1667em\lower.7ex\hbox{E}\kern-.125emX}}
\usepackage{listings}

\definecolor{codegreen}{rgb}{0,0.6,0}
\definecolor{codegray}{rgb}{0.5,0.5,0.5}
\definecolor{codepurple}{rgb}{0.58,0,0.82}
\definecolor{backcolour}{rgb}{0.95,0.95,0.92}

\lstdefinestyle{mystyle}{
    backgroundcolor=\color{backcolour},   
    commentstyle=\color{codegreen},
    keywordstyle=\color{magenta},
    numberstyle=\tiny\color{codegray},
    stringstyle=\color{codepurple},
    basicstyle=\ttfamily\footnotesize,
    breakatwhitespace=false,         
    breaklines=true,                 
    captionpos=b,                    
    keepspaces=true,                 
    numbers=none,                    
    numbersep=5pt,                  
    showspaces=false,                
    showstringspaces=false,
    showtabs=false,                  
    tabsize=2
}

\newcommand{\code}[1]{\texttt{#1}}

\usepackage{hyperref}
\begin{document}

\title{Large-Scale Application of Fault Injection into PyTorch Models - an Extension to PyTorchFI for Validation Efficiency}

\author{
\IEEEauthorblockN{Ralf Gräfe\IEEEauthorrefmark{1}, Qutub Syed Sha\IEEEauthorrefmark{1}\IEEEauthorrefmark{2}, Florian Geissler\IEEEauthorrefmark{1}, Michael Paulitsch (\textit{Senior Member, IEEE})\IEEEauthorrefmark{1}\\
}
\IEEEauthorblockA{\IEEEauthorrefmark{1} Intel Labs, Germany\\}
\IEEEauthorblockA{\IEEEauthorrefmark{2} Technical University of Munich, Germany}
}

\maketitle

\begin{abstract}
Transient or permanent faults in hardware can render the output of Neural Networks (NN) incorrect without user-specific traces of the error, i.e. silent data errors (SDE). On the other hand, modern NNs also possess an inherent redundancy that can tolerate specific faults. To establish a safety case, it is necessary to distinguish and quantify both types of corruptions. 

To study the effects of hardware (HW) faults on software (SW) in general and NN models in particular, several fault injection (FI) methods have been established in recent years. Current FI methods focus on the methodology of injecting faults but often fall short of accounting for large-scale FI tests, where many fault locations based on a particular fault model need to be analyzed in a short time. Results need to be concise, repeatable, and comparable.

To address these requirements and enable fault injection as the default component in a machine learning development cycle, we introduce a novel fault injection framework called PyTorchALFI (Application Level Fault Injection for PyTorch) based on PyTorchFI. PyTorchALFI provides an efficient way to define randomly generated and reusable sets of faults to inject into PyTorch models, defines complex test scenarios, enhances data sets, and generates test KPIs while tightly coupling fault-free, faulty, and modified NN. In this paper, we provide details about the definition of test scenarios, software architecture, and several examples of how to use the new framework to apply iterative changes in fault location and number, compare different model modifications, and analyze test results.

\end{abstract}


\section{Introduction}

Ongoing shrinkage of transistor sizes in computer silicon and accompanying lower operating voltage leads to increased HW faults like bit flips during operation. This has high relevance in safety-critical applications like autonomous driving. 

The research by \cite{IBRAHIM2020113969} comprehensively reviews probable soft errors that may arise in deep neural network (DNN) accelerators. The study specifically highlights transient faults that occur in the form of bit flips, which are a crucial concern for the model's reliability and resilience. \cite{Li2017} and \cite{10.1007/978-3-319-99130-6_14                       } show evidence that single transient bit-flips can lead to perturbations of DNN at the application level. \cite{Torres-Huitzil2017} gives an overview of the types of hardware faults relevant to the execution of CNNs. Recent studies by \cite{qutub2022hardware} and \cite{geissler2021towards} offer additional insights into the susceptibility of object detection and image classification models to hardware faults. These studies underscore the importance of detecting and mitigating faults in DNNs to maintain model accuracy and reliability.

Determining the effect of HW faults on DNN is not straightforward because they have an unspecific redundancy and robustness against single faults in the computed path. However, a certain percentage of faults do cause output corruption in the DNN. In the following we refer to this effect as Silent Data Error (SDE). It is, therefore, important to analyze the sensitivity of the DNNs to the typical forms of hardware-related faults. \cite{Li2017} also shows that the number of parameters in a DNN model determines if a hardware fault leads to a critical failure depending on where in the network or input data the fault occurs. HW faults can be efficiently modelled as bit flips on the application level. A bit flip can affect different bit positions of a value where the most signification bits, e.g. exponent bits in floating point numbers, have the highest impact, different quantization could be used, and weights or neurons could be hit. Furthermore, different layers of a DNN can be more or less susceptible, or even different input classes can react differently to faults. Quantifying those effects on DNN enables developers to concentrate on the most vulnerable parts when making DNN more resilient and also provides vulnerability factors for formulating safety cases.

To support this analysis, an established method is to perform fault injection at different HW/SW stack levels. Hardware stack fault simulation offers greater precision but is impractical for extensive fault injections in modern models with over 10 million parameters. Also, the data type amplifies the number of vulnerable bits to be tested. For example, a 16-bit model with over 10 million parameters will result in 160 million vulnerable bits being tested against hardware faults, further decreasing testing efficiency. To make fault injection an efficient part of a continuous product development cycle, it must support large-scale fault injections with accurate logs to reproduce fault locations and compare results between fault-free, faulty, and improved hardened networks. It also needs to support the generation of performance metrics and match them with input data and fault locations. 

This paper describes a new fault injection framework called PyTorchALFI (Application Level Fault Injection for PyTorch) which extends PyTorchFI \cite{Mahmoud2020} to make it better suited for large-scale fault injection campaigns. PyTorchALFI adds the following features to the original PyTorchFI:
\begin{itemize}
	\item Simple definition of fault injection metadata: control total number of faults, type, and location through a configuration file and make fault definition and test parameters available for subsequent experiments.
	\item Definition of complex test scenarios: easy iteration over different aspects of fault injection tests like location, different number of faults and mapping corresponding test results.
	\item Tight integration of fault-free, faulty, and enhanced models: Enables synchronized inference and results in logging of separate DNN instances, allowing for comparison of fault effects on the original, fault-injected model and the modified mitigation model. This analysis can be done at a granular level of a single fault location and input data point or at a statistical level.
	\item Data set enrichment: existing data sets are wrapped to provide additional metadata to enable later reproduction of fault conditions down to a single data item (e.g., a single image)
	\item KPI generation: commonly used and new KPIs are automatically calculated at the end of test runs.
\end{itemize}

In the remainder of this document, we give additional details on relevant aspects of PyTorch and PyTorchFI in the background section. We then explain how the SW architecture follows large-scale repeatable fault injection requirements. The evaluation section provides examples and reasoning to support how PyTorchALFI enhances application-level fault injection efficiency and explains the process of generating test results. The conclusion section finally provides a summary of the proposed approach, results, and future work.
The tool is available as source code here: \url{https://github.com/IntelLabs/PyTorchALFI}.

\section{Background}
PyTorch is a popular Python framework for training and running neural networks. The framework offers flexibility in redefining the network's structure and computational path, enabling analysis and modification of each layer's input and output values during training and inference. This can happen by inserting additional processing steps or layers in the forward function or attaching hooks, i.e., callback functions, to selected model parts to modify input or output values on the fly. The PyTorch hook functionality can be used for fault injection and is indeed used for this purpose in PyTorchFI.

One of the hook function parameters is the output of a specific layer's MAC operation. The output values are modified in place. A fault injection routine can thus change one or more values in the $output$ tensor. For instance, it can simulate faults in either the hardware component performing the MAC operation or the register storing the output values before forwarding them to an activation function. To clarify, hooks are used for fault injection in neurons, since the values of the tensor position that are to be corrupted are only determined during run time. Fault injections into weights don't have to use hooks, because weights are defined before the inference run. PyTorchFI, the fault injection framework developed by the RSim Research Group~\cite{RSim}, applies the above-mentioned methods to inject faults into neurons and weights. Its source code is available here: \url{https://github.com/PyTorchfi/PyTorchfi}. A branched-off version of PyTorchFI is used as the core of PyTorchALFI. 

Integrating a model for vulnerability assessment can be challenging, as the existing framework is not well suited for handling large-scale fault injections. To address these issues, we propose a new framework - PyTorchALFI.

\section{Related Work} 

Faults can be injected directly into the HW, HW simulations, or the application layer. 
Examples of HW level FI are RIFLE \cite{Madeira1994} or FIST \cite{259894}, the first using pin-level fault injection and the latter using ion radiation. Simulation-based FI using languages like \code{VHDL} or \code{Verilog} is implemented in MEFISTO \cite{Jenn1994}, DEPEND \cite{Goswami1997} or VERIFY \cite{Sieh1997}. Further techniques known as MUTANT and SABOTEUR are described in \cite{Na2011}, and \cite{Baraza2005}.  Software Level FI is demonstrated in FERRARI \cite{Kanawati1995}, XCEPTION \cite{Carreira1995XceptionSF}, LLFI, and PINFI \cite{Wei2014}. This paper focuses on application-level fault injection where either bit(s) are flipped or original values are randomly changed. So far, only a few FI tools that specialize in neural networks at the application level exist.  Two examples of the frameworks TensorFlow and PyTorch are TensorFI \cite{Li2018} and PyTorchFI \cite{Mahmoud2020}.

\section{Methodology}
\subsection{Requirements}
Our primary objective was to efficiently define, execute, and analyze large-scale fault injections, ensure the persistence and re-usability of utilized faults, and efficiently extract test results. There is a further requirement to systematically move the focus of the fault injection to specific layers of the CNN. Storing and reusing fault locations is essential to ensure the comparability and reproducibility of the researcher's experiments.  

We reduced these high-level requirements to the following needs. The different components of a neural network that should be addressable by fault injection are:

\begin{itemize}
	\item types of layers
	\item location of a layer within the model
	\item location of a fault within a layer
	\item injection into neurons or weights
	\item numeric type used and bit position within this numeric type
    \item fault model
\end{itemize}

It should also be possible to set the number of faults that are applied to the input, e.g., a single image in the case of CNN, at the same time. Finally, for fault injection into weights, it is essential to decide the scope of a (weight) fault; this means defining whether a specific fault is applied to a single image, a batch of images, or even a whole epoch involving a complete test data set. Finally, the fault model should support both transient and permanent faults.

\subsection{Architectural design} \label{subsec:architectural-design}
\begin{figure}[h]
\centering
\includegraphics{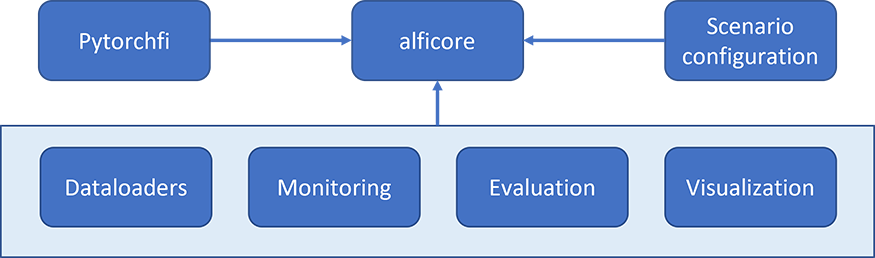}
\caption{Software architecture of PyTorchALFI}
\label{fig:swarch}
\end{figure}

Figure~\ref{fig:swarch} shows the software architecture for PyTorchALFI. The following components extend the original PyTorchFI solution. The tool's core is the alficore component, which provides a test class that integrates all functionalities. The specifics of the fault injection campaign, such as the fault model and the number of injected faults, will be configured in the scenario configuration component. The enhanced PyTorchFI component will take care of the fault injection execution. One of its core features is the use of PyTorch hooks. This allows the modification of neuron values in place. 

In addition to its core functionalities, alficore offers various optional features in the form of user-friendly data loader wrappers, monitoring capabilities (enabling the detection of NaN or Inf values and facilitating the integration of custom monitoring), evaluation functions (allowing the calculation of SDE or Detected and Uncorrected Errors (DUE) rates), and visualization tools, permitting the plotting of critical results, currently limited to object detection.

A fault injection campaign starts with an existing application that initiates a pre-trained PyTorch model and performs an inference run with input from a data set. This application is then augmented with PyTorchALFI function calls to perform fault injections into weights or neurons.

Once PyTorchALFI has been added to the project folder, the parameter definitions are established in a distinct configuration scenario yml file called default.yml. After verification, these parameters are accessible during run time and can be altered as needed. For instance, iterative modification of the number of faults per inference can be performed, allowing for customization of the number of faults per image, batch, or epoch. Faults can be inserted in weights or neurons in a PyTorch model. It is possible to define the type of layer where fault injection happens. Supported layer types are conv2d, conv3d, and Linear (fully connected). The fault injections can be limited to specific layer numbers or a range of layer numbers. Furthermore, the random selection of layers can be weighted to increase the likelihood of selecting more significant layers. 

Modifications can be made to either numbers or specific bits of neurons/weights. In the former, random values are drawn from a specified min-max range of values. The latter identifies a range of bits from which the flipped bit is applied. The number of faults per image can be expressed as a fixed integer or a distribution. In the latter case, the distribution represents a fraction of the total number of weights or neurons.

All faults are generated as a matrix before the inference run to enhance the explainability of faults. Each column in the matrix contains a single fault. Fault definitions comprise the fault location and value. The location parameters vary slightly between weight and neuron fault injections. Regarding neuron fault injection, the various rows in the matrix hold significance, as illustrated in Table ~\ref{tab:faultloc}.
\begin{table}[h]
	\centering
		\noindent
		\begin{tabular}{|lll|}
		\hline
		line number & ID & Description \\
		\hline
		\hline
		1 & Batch & number of images within a batch \\
		\hline
		2 & Layer & n\textsuperscript{th} layer out of all available layers \\
		\hline
		3 & Channel & n\textsuperscript{th} channel out of all available channels \\
		\hline
		4 & Depth & additional index for conv3d layers \\
		\hline
		5 & Height & y position in input \\
		\hline
		6 & Width & x position in input \\
		\hline
		7 & Value & either a number or the index of bit position \\
		\hline
		\end{tabular}
	\caption{Fault definition parameters for neuron fault injection}
	\label{tab:faultloc}
\end{table}

Weight fault injection adheres to similar definitions as presented in Table ~\ref{tab:faultloc}, with minor differences. The first row denotes the layer index, and the second and third rows specify the weight's output and input channel, respectively. The number of columns (i.e., total faults) in the fault matrix is determined based on the data set size and the fault injection policy, which can be set per image, batch, or epoch. After generating the faults, the fault matrix is stored as a binary file. Consequently, the identical set of faults can be utilized across various experiments to evaluate the impact of model modifications on fault mitigation. A second binary file is generated after the fault injection experiment. This file comprises information on the fault locations and the original and altered values of the neuron/weight before and after the fault injection run. Additionally, it includes other monitored values, such as Inf and NaN, associated with each inference. PyTorchALFI saves all experiment parameters in a yml file format, which can be used to replicate an experiment for future purposes accurately. 

PyTorchALFI provides an iterator that returns the original model with different faults applied at each call during the inference run. The number and position of these faults are taken successively from the fault matrix. The initial PyTorch application determines the model used for fault injection, the data set size, and the number of epochs. The developer is flexible in how to use the faulty models created by PyTorchALFI in the application.

\section{Using PyTorchALFI to perform experiments and store results} \label{usage}

This section gives some examples of PyTorchALFI runs starting with a principle integration into one's code, including parameters settings and results given the 
particular test goals in the following list.

\begin{enumerate}
	\item \emph{inject faults at random positions throughout the network} -- to determine the probability of failure in the model output in the presence of hardware faults producing bit flips
	\item \emph{iterate through a model} to verify if certain components are more \emph{sensitive to faults} than others
	\begin{enumerate}
            \item \emph{iterate through single or groups of layers} to determine which layers are more susceptible to errors.
		\item successively \emph{increase the number of concurrent faults per image} -- to verify how robust a model is and how many faults it can tolerate.
		\item \emph{switch} between neuron and weight fault injection -- to determine if a mitigation strategy (if applied) is equally effective for neuron or weight faults.
		\item \emph{change the bit flip  position} for either weight or neuron faults -- to verify which bit positions with a particular fault model are likely to produce failures in the output. 
	\end{enumerate}
\end{enumerate}

PyTorchALFI can thus be used to implement the following use cases to PyTorch models.

\begin{itemize}
	\item Finding the most vulnerable components in a NN
	\item Evaluating the vulnerability of different numeric types
	\item Comparing the robustness of different types of NN
	\item Verifying the efficiency of mitigation strategies against faults or attacks
	\item Compare the robustness of NN between the original model and a pruned version
	\item Finding the maximum amount of faults per image where the accuracy of the output of the NN stays within acceptable boundaries
\end{itemize}

Independent of the use case, PyTorchALFI needs to be integrated with the existing NN project. There is a high-level and low-level method to do that. The integration methods are described first. After that, the implementation of use cases is described.

\subsection{Integration into own project - low level}

In the low-level method, only fault-injected models and their output is provided. The developer is free to use the raw results as needed. The pseudo-code listing ~\ref{code:low-integration} shows how to integrate PyTorchALFI into existing code using the low-level integration mode.

\begin{lstlisting}[language=Python, label=code:low-integration, caption=Pseudo code to integration PyTorchALFI into custom code]
from PyTorchfiwrapper.wrapper.ptfiwrap import ptfiwrap

# Initiate the wrapper with the trained baseline model 
net=orig_model
wrapper = ptfiwrap(model=net)

# get an iterator over faulty models
fault_iter = wrapper.get_fimodel_iter()

for [loop through epochs <@\textcolor{black}{and data set}@>]
		CORRUPTED_MODEL = next(fault_iter)
		orig_output = orig_model(input)
		Corrupted_output = CORRUPTED_MODEL(input)
		# store and analyze Corrupted_output in 
		# comparison to orig_output

\end{lstlisting}

The original model is handed over to the \code{ptfiwrap} object. \code{fault\_iter} then provides a complete NN model with one or more new faults injected. Depending on the specific fault model, the developer can manipulate the frequency at which faults are altered by positioning the iterator at different locations within the program loop. For example, faults can be changed for each image, batch or once per epoch.
The code expects the file \code{default.yml} inside folder \code{scenarios}. Details are given in the following sub-sections.

Regarding the low-level integration, PyTorchALFI stores two binary files and a scenario yml file.

\subsection{Integration into own project - high level}
High-level integration of PyTorchALFI addresses image classification and object detection networks using CNN by providing relevant result files. A separate class is provided for each case, \code{test\_error\_models\_imgclass.py} and \code{test\_error\_models\_objdet.py} respectively. These classes encapsulate the generation of wrapper instances that iteratively create faulty versions of the model under test and run it with batches of input images from the chosen data set. Listing ~\ref{code:high-integration}  provides an example of high-level integration into one's code using yolov3.

\begin{lstlisting}[language=Python, label=code:high-integration,caption=Pseudo code to integration PyTorchALFI into custom code]
from PyTorchfiwrapper.models.yolov3.darknet import Darknet
from PyTorchfiwrapper.wrapper.test_error_models_objdet import TestErrorModels_ObjDet

# Initiate the model 
yolov3 = Darknet("PyTorchfiwrapper/models/yolov3/config/yolov3.cfg").to(device)
yolov3.load_darknet_weights("PyTorchfiwrapper/models/yolov3/weights/yolov3.weights")
model = yolov3.eval().to(device)

model_ErrorModel = TestErrorModels_ObjDet(model=model, resil_model=None, model_name=model_name, data set=data set_name, config_location=yml_file, dl_shuffle=False, device=device)
				
model_ErrorModel.test_rand_ObjDet_SBFs_inj(fault_file='', num_faults=nr_faults, inj_policy='per_image')

\end{lstlisting}

This simple integration reads all configuration information from the \code{config\_location} and initiates the data set given in the \code{data set} parameter. Several data sets are prepared for use with these classes. They provide data in the form of lists: \code{[dict\_img1\{\}, dict\_img2(), dict\_img3()] -> dict\_img1 = \{'image':image,'image\_id':id, 'height':height, 'width':width ...\}} to conserve the image locations for later results analysis. The primary purpose of these higher-level classes is the easy and concise preparation of test results. Each fault injection run produces up to 3 sets of outputs. \textbf{One} is a meta-file that contains information about all parameters used to run the fault injection. This includes the used model, the location of its implementation, the data set used, and the corresponding data loader, as well as any test parameters modified during the run in the form of a yml file. \textbf{The second} is a set of binary files of all faults and bit position changes (from $0$ $\rightarrow$ $1$ or vice-versa) that were applied in the test and can be reloaded using a configuration parameter. \textbf{The third} are default vulnerability results in the form of CSV or JSON files (CSV in the case of classification and JSON in the case of object detection networks). 

The results files allow a detailed analysis of the effects of applied faults down to the position of bits flipped and their flip direction for single images.	

\subsection{Random positions throughout the network}
\label{sec:random}
This section describes the necessary configurations to apply faults to random locations throughout a network. This relates to the test goal listed in Section \ref{usage} item 1.

The initial configuration for each fault injection run happens in file default.yml. The file itself contains documentation on each parameter. The wrapper is designed to pre-generate all faults for the complete run. For any run, the total number of different faults that are needed is determined by the three variables $data set\_size\ (a)$, $num\_runs\ (b)$ and $max\_faults\_per\_image\ (c)$. $data set\_size$ states how many images are contained in the used data set or how large the data subset will be used for the test. $num\_runs$ states how often the data set will be used (epochs), and $max\_faults\_per\_image$ states how many different faults should be injected for the same image or rather while processing a single image. The number of faults $n$ to prepare is determined by calculating the product of all three values $n = a \cdot b \cdot c$.
Two types of faults can be injected: random numeric values or bit flips at given bit positions. For realistic simulation of hardware faults, bit flips should be used. The range of bits used for input values and weights and the content of bit positions that can be randomly selected must be specified. To allow all possible values in a 32-bit number, this value must be set to, e.g. $rnd\_bit\_range: [0,31]$. 

Apart from bit position, the location of layer number and values for height, width, and channel inside a layer are randomly selected. In addition, each layer's relative size can be considered to determine the likelihood it is drawn. The relative size of each layer is calculated separately for weights and neurons by summing the quotient of the number of elements per layer divided by the total number of elements in the model.

\begin{equation}
	F_i = \frac{\prod_{j=1}^{m}d_{ij}}{\sum_{i=1}^{n}\prod_{j=1}^{m}d_{ij}}
\end{equation}
 
Where $F_i$ is the weight factor for layer $i$ and $d_{ij}$ are the sizes of the different dimensions of the layers tensor.
The layer types (Conv2D, Conv3D, or fully connected) can be specified to define fault locations further.

The location of faults is further determined by specifying if faults are injected in the neurons or the weights. Neurons and weights cannot be selected in the same fault injection run.

\subsection{Iterate through a model}
All usage models listed in Section \ref{usage} test goal item 2 can be implemented in this way.
Extending the concept described in the previous section, where parameters are set only once at the beginning of the fault injection run, for iterations of any sort, the original parameter set is accessible and can be modified during run-time. 
When iterating through layers, the start layer is set to an initial value, e.g., 1. 
After that, the data set is iterated over once. After that, the parameter can be reset to the following layer number and rewritten using the functions \code{wrapper.get\_scenario()} and \code{wrapper.set\_scenario()}.
This also generates a new set of faults for the whole data set. 
A similar approach can be applied for iterating through faults per image or bit position for bit flips. 
In addition, a change between neuron and weight faults is equally possible. This method allows the efficient setup of fault injection scenarios without manual reconfiguration after each set of fault parameters.
\subsection{Data loader wrapper}
Dataloaders are enhanced to allow capturing additional information during fault injection runs. The minimal information stored about each image is directory+filename, height, width, and image id. This can be extended to individual needs. For consistent handling of different data sets, each dataset is first brought into a JSON format as used in the COCO data set \cite{coco}. Facebook's detectron2 project inspires PyTorchALFI's data loader \cite{wu2019detectron2}. It builds on the user's existing data loader. Currently, PyTorchALFI supports COCO-based Average-Precision metric variant (AP) evaluation.
For instance, Intersection over Union (IoU), average precision (AP), and average recall (AR) are computed using COCO's defined metrics.

\subsection{Result evaluation}
\begin{figure}
\centering
\begin{subfigure}[b]{0.5\textwidth}
\includegraphics[width=0.98\linewidth]{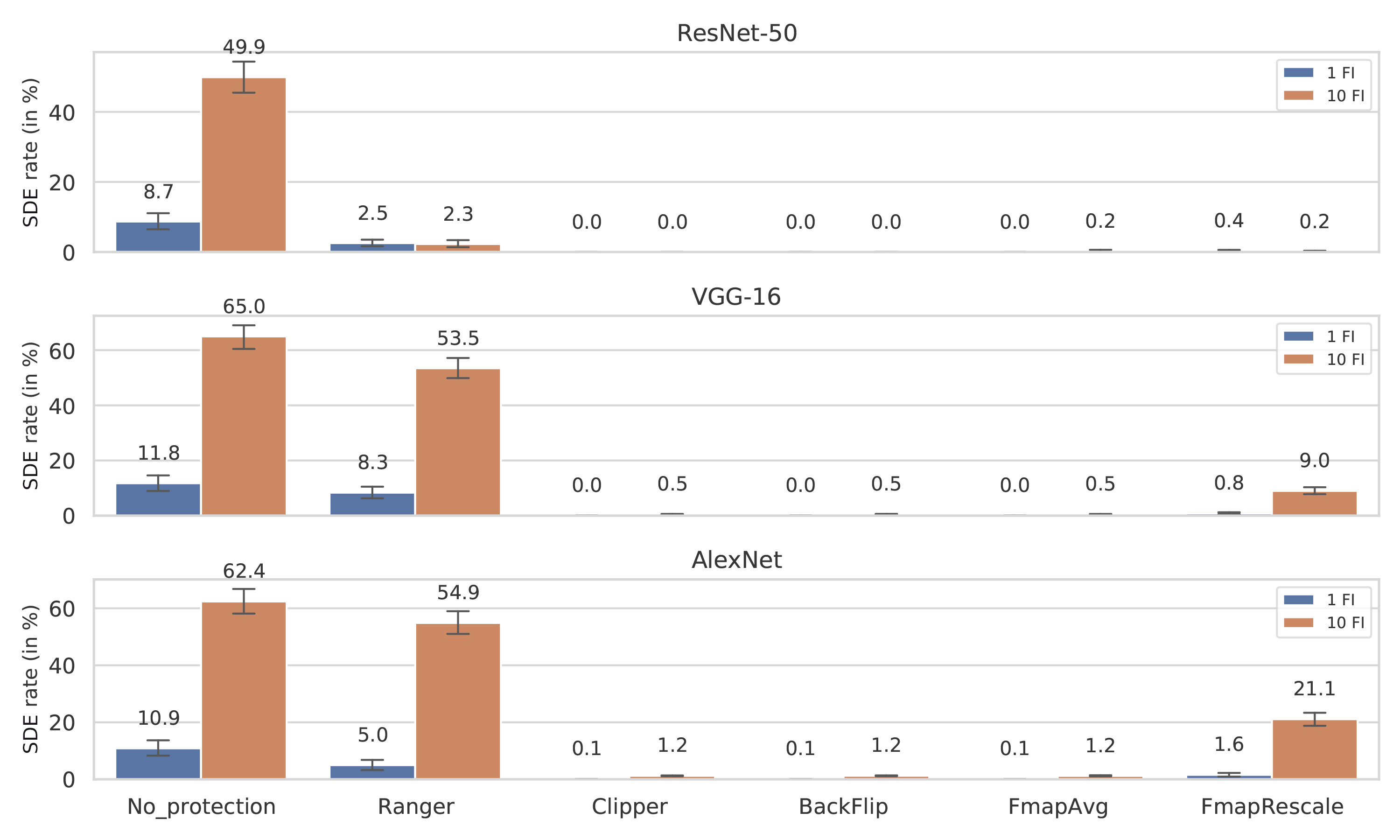}
\caption{SDE rates for image classification models: Our previous work cited in \cite{geissler2021towards}, forms the basis for these results}
\label{fig:class_eval} 
\end{subfigure}
\begin{subfigure}[b]{0.5\textwidth}
\includegraphics[width=0.98\linewidth]{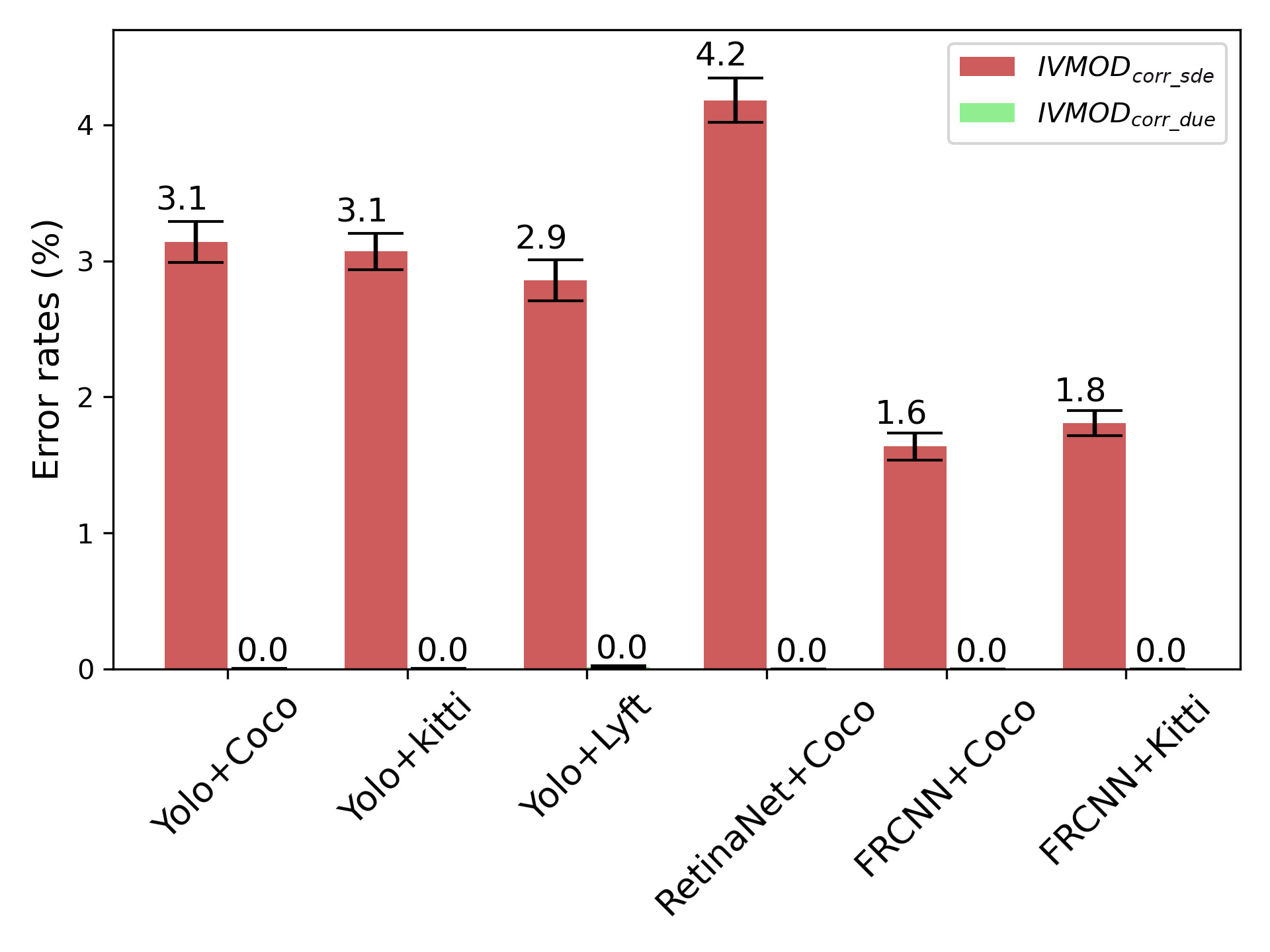}
\caption{SDE rates for object detection models: Our previous work cited in \cite{qutub2022hardware}, forms the basis for these results}
\label{fig:obj_eval} 
\end{subfigure}
\caption{SDE rates for weight fault injection}
\label{fig:SDE_eval}
\end{figure}

PyTorchALFI stores result from fault injection experiments in convenient CSV (for classification models) and JSON (for object detection models) formats. The SDE results presented in Figure ~\ref{fig:SDE_eval} were obtained from previous research conducted using the PyTorchALFI tool for conducting fault injection experiments, as described in \cite{geissler2021towards} and \cite{qutub2022hardware}.
The focus lies on classification and object detection CNN, which have distinct KPIs. 

\subsubsection{Classification CNN}	
	Generally, the output for classification CNN focuses on the top 5 results, and hence PyTorchALFI, by default, stores top-5 classes and probabilities in addition to the ground truth label. 
	Furthermore, the CSV file stores fault positions defined by layer, channel, height, width, and bit position that was flipped. This can be easily extended to top-K classes.
	In case of multiple faults per image, these fault positions are listed. PyTorchALFI stores three sets of outputs. 
$a)$ meta-files - scenario.yml and the data loader’s absolute file path,
$b)$ fault location binary files,
$c)$ models output.
The model's fault files contain the same faults fed during the fault injection run. 
In addition, model fault files include bit flip direction and original and changed values of neurons/weights. 
The model's outputs are the final top-5 output of the models (original model and model applied mitigation). In addition, outputs without fault injections are stored in a separate CSV file. This raw basic information is further processed to quantify the vulnerability. For example, the classification model's silent data error (SDE) can be quantified easily using the third set of model files.
For example, Figure~\ref{fig:class_eval} studies the SDE of ResNet-50, VGG-16, and AlexNet. The faults were injected at weight level only on exponential bits. Using the first set of outputs binary files, bit-wise and layer-wise, SDE information was easily extracted. As shown in  Figure~\ref{fig:class_eval}, VGG-16 without protection has an 11.8\% vulnerability when injected with a single fault per image inference. With the PyTorchALFI tool's scalability, vulnerability analysis with different protections (e.g. Ranger/Clipper \cite{geissler2021towards}) can be easily extracted for various models without additional adaptation.

\subsubsection{Object detection CNN} 
\begin{figure}[h]
\centering
\includegraphics[width=\linewidth]{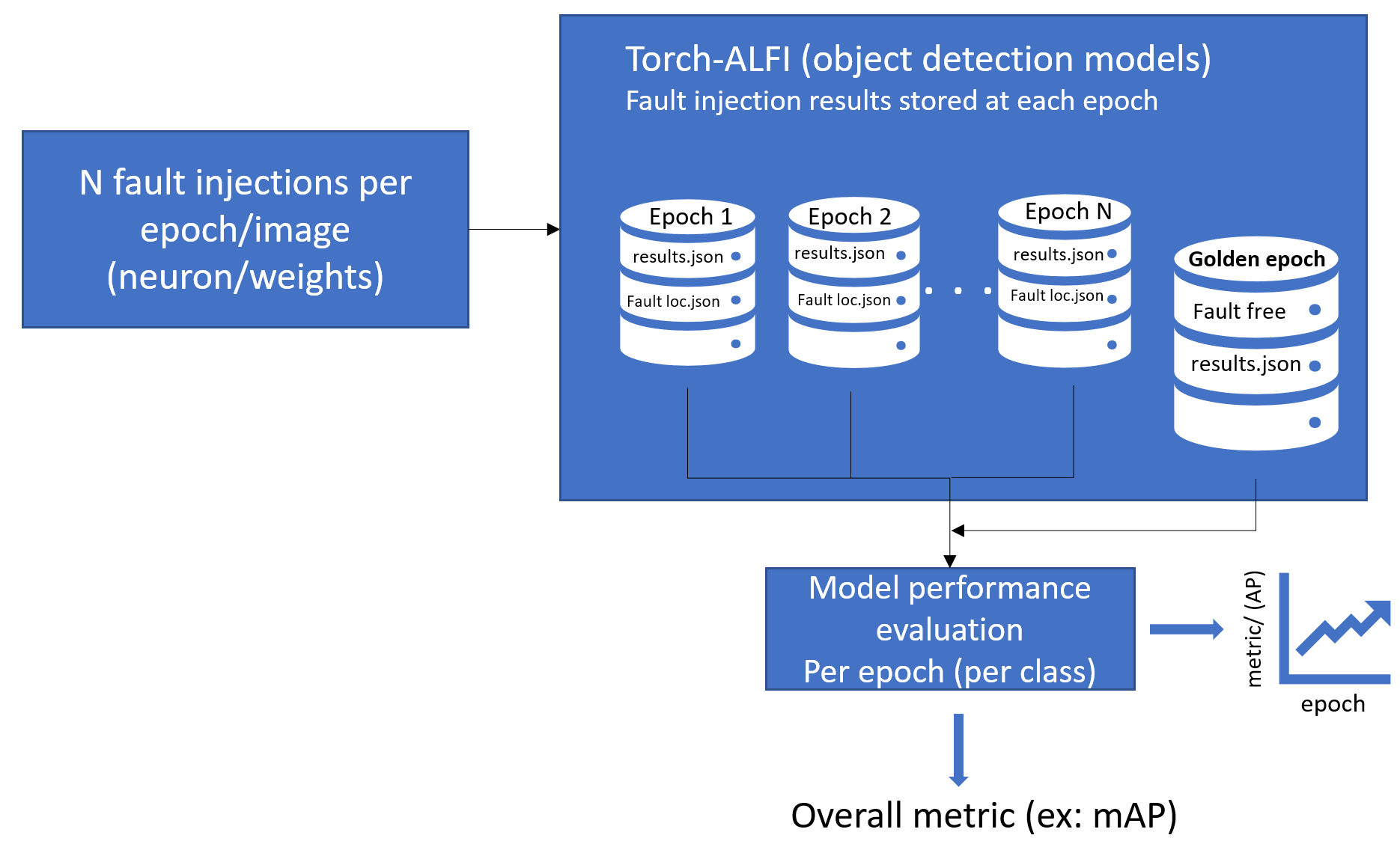}
\caption{PyTorchALFI's Object detection submodule}
\label{fig: ptfiwrap_objdet}
\end{figure}

PyTorchALFI, in this case, also generates three sets of output files: 
$a)$ Ground truth and Meta-files - the ground-truth JSON file, a list of annotations for each image bounding boxes and class ID, as well as scenario.yml storing all the run-time parameters used, 
$b)$ intermediate result lists containing predicted classes, scores, and bounding box location per object and 
$c)$ mAP values.
Object detection models are rather complex compared to classification models. Hence, intermediate results of an entire data set are exposed with one set of faults, each stored in separate JSON files as shown in Figure~\ref{fig: ptfiwrap_objdet}. During the post-processing stage of these results, the stored JSON files are used to analyze the impact of the fault injection. These JSON files are 
processed using metrics like COCO-API \cite{coco} and IVMOD \cite{qutub2022hardware} to estimate the object detection model's vulnerability. For example, $IVMOD_{SDE}$ is shown in Figure~\ref{fig:obj_eval} for the models YoloV3, Retina-Net and Faster-RCNN (FRCNN) trained on multiple datasets.
As an example, when injected with a single fault per image inference, RetinaNet trained on CoCo has a vulnerability of 4.2\% in producing incorrect detections. Moreover, it has a low probability ($<10^{-2}$) of generating NaN/Inf values - $IVMOD_{DUE}$.

\subsection{Extensibility}
PyTorchALFI has been designed with further extensibility, enabling the easy addition of more supported model types or result metrics. 
The tool is designed to easily incorporate new custom trainable layers not native to Pytorch by adding the custom layer's type in the $verify\_layer$ function within the $errormodels.py$ file. This feature provides flexibility in expanding the tool's capabilities to support various model types. New signals at intermediate layers can also be efficiently monitored by including their respective monitoring functions in the $attach\_hooks$ functions in the $TestErrorModels\_X$ classes, enabling a more comprehensive analysis of the model's behaviour during fault injection. This modularity and flexibility make the tool suitable for researchers to adapt and extend for their specific use cases.
Even the actual fault injector can be replaced or improved in the future. First, tests have been performed to integrate a fault injection method that relies on low-level ML library primitives to provide a more realistic fault behaviour based on faults in specific HW units that perform the MAC operations in Convolutional Neural Networks. This work is still ongoing and not finalized. The mentioned fault injection method is described in \cite{Omland2022}.

\section{Conclusion}
We presented the PyTorchALFI framework, focusing on efficiently incorporating fault injection into the regular SW development cycle, especially for safety-critical NN applications built with PyTorch. We identified the main requirements for efficient definition, execution and result analysis of large-scale fault injections, persistence and re-usability of used faults, and efficient extraction of test results. Based on that, we developed our architecture by enhancing the existing PyTorchFI tool with an easy definition of fault injection metadata, a detailed description of complex test scenarios, the tight coupling of fault-free, faulty and improved models, data set enhancement, and the generation of KPI. We finally showed how our framework could be integrated into existing PyTorch projects and how it is configured for different use cases.

The tool is available as source code here: \url{https://github.com/IntelLabs/PyTorchALFI}.

\section{Acknowledgment}
\begin{flushleft}
\begin{table}[ht]
\begin{tabular}{@{}b{0.1\textwidth}b{0.35\textwidth}@{}}
\includegraphics[width=1.5cm]{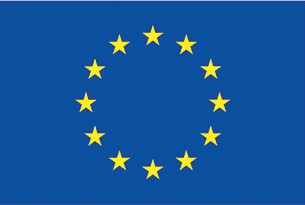}&This project has received funding from the European Union’s Horizon 2020 research and innovation program under grant agreement No 956123.
\end{tabular}
\end{table}
\end{flushleft}

\bibliographystyle{ieeetran}
\bibliography{ptfiwrap}
\end{document}